%% file: main.tex
\documentclass[11pt]{article}
\usepackage[utf8]{inputenc}
\usepackage{graphicx,amsmath,amsfonts,amssymb,fullpage}

\usepackage[T1]{fontenc}

\usepackage{url}
\usepackage{booktabs}
\usepackage{nicefrac}
\usepackage{microtype}
\usepackage{xcolor}

\usepackage{multirow}
\usepackage{graphicx}
\usepackage{wrapfig}
\usepackage{amssymb}
\usepackage{amsmath}
\usepackage{subfig}
\RequirePackage{hyperref}

\definecolor{darkblue}{rgb}{0, 0, 0.5}
\hypersetup{colorlinks=true, citecolor=darkblue, linkcolor=darkblue, urlcolor=darkblue}

\usepackage[most]{tcolorbox}
\tcbuselibrary{listingsutf8}

\newtcolorbox[
  auto counter,
]{mybox}[2][]{
  enhanced, 
    breakable,
    skin first=enhanced,
    skin middle=enhanced,
    skin last=enhanced,  
  coltitle=black, 
  fonttitle=\bfseries,
  attach title to upper={\ },
  title=Box \thetcbcounter: #2, #1
}

\newtcolorbox[
  auto counter,
]{myboxNonumber}[2][]{
  enhanced, 
  coltitle=black, 
  fonttitle=\bfseries,
  attach title to upper={\ },
  title=#2, #1
}

\input{math_commands}

\usepackage[numbers, compress, super]{natbib}

\title{\textbf{Overcoming classic challenges for artificial neural networks by providing incentives and practice}}

\author{Kazuki Irie$^{1}$ ~ Brenden M.~Lake$^{2}$\\
$^{1}$Department of Psychology, Harvard University, Cambridge, MA, USA \\ $^{2}$Departments of Computer Science and Psychology\\ Princeton University, Princeton, NJ, USA\\
\texttt{kirie@g.harvard.edu}, \texttt{brenden@princeton.edu}}
\date{}

\usepackage{fancyhdr}
\fancypagestyle{plain}{%
  \fancyhf{}%
  \fancyhead[L]{\itshape In press at Nature Machine Intelligence}%
}

\begin{document}

\maketitle

\begin{abstract}
Since the earliest proposals for artificial neural network (ANN) models of the mind and brain, critics have pointed out key weaknesses in these models compared to human cognitive abilities.
Here we review recent work that uses metalearning to overcome several classic challenges, which we characterize as addressing the Problem of Incentive and Practice---that is, providing machines with both incentives to improve specific skills and opportunities to practice those skills.
This explicit optimization contrasts with more conventional approaches that hope the desired behaviour will emerge through optimizing related but different objectives.
We review applications of this principle to addressing four classic challenges for ANNs: systematic generalization, catastrophic forgetting, few-shot learning and multi-step reasoning. 
We also discuss how large language models incorporate key aspects of this metalearning framework (namely, sequence prediction with feedback trained on diverse data), which helps to explain some of their successes on these classic challenges.
Finally, we discuss the prospects for understanding aspects of human development through this framework, and whether natural environments provide the right incentives and practice for learning how to make challenging generalizations.

\end{abstract}

\section{Introduction}
\label{sec:intro}
\begin{figure*}[t!]
    \begin{center}
        \includegraphics[width=0.8\columnwidth]{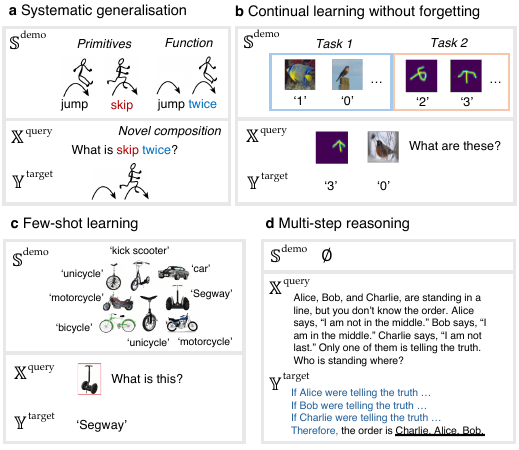}
        \caption{Illustrations of four examples of cognitive challenges for artificial neural networks (ANNs).
        For each problem, symbols $\mathbb{S}^\text{demo}$, $\mathbb{X}^\text{query}$, and $\mathbb{Y}^\text{target}$ indicate the task demonstrations, query, and  target output, respectively (using the corresponding notations of the PIP framework; Fig.~\ref{fig:general_framework}).
        \textbf{a:} Systematicity deals with compositional generalization; by learning how to ``jump'', ``skip'', and ``jump twice'', we can infer how to ``skip twice''.
        \textbf{b:} Catastrophic forgetting refers to ANNs' failure to remember previously learned skills after learning to solve another task; in this example, Task 1 consists in classifying animal images, while Task 2 is classification of some characters/signs.
        \textbf{c:} Few-shot learning is showcased in how people can learn about a ``Segway'' from one example and then recognize other examples of the concept.
        \textbf{d:} Multi-step reasoning involves solving a problem by decomposing it into a series of logical steps.
        }
        \label{fig:classic_challenges}
    \end{center}
\end{figure*}

As long as there has been artificial intelligence (AI), AI has been compared to its natural counterpart, human intelligence.
Artificial neural networks
(ANNs; also known as \textit{connectionist models} and parallel distributed processing, see Glossary in Box \ref{box:glossary}) \citep{mcculloch1943logical, rosenblatt1958,rumelhart1986general,mcclelland1986pdp} have long been criticized for certain failures relative to human intelligence (see Fig.~\ref{fig:classic_challenges}), leading to debates about their capabilities that span cognitive science, AI, and related fields. For example, critics have argued that ANNs lack \textit{systematic compositionality}\citep{fodor1988connectionism,marcus1998rethinking,lake2017generalization,greff2020binding}---the algebraic ability to understand or produce novel combinations of simpler components, or show \textit{catastrophic forgetting during continual learning} \citep{mccloskey1989catastrophic,ratcliff1990connectionist,french1991using,french1999catastrophic}---the ability to learn new skills without forgetting old skills. Critics have also argued that ANNs are excessively data hungry \citep{geman1992neural} and  struggle with \textit{few-shot learning}---the ability to learn a new concept from just one or a few examples\citep{MillerMV00,FeiFei03,lake2011one,lake2015human}; and that ANNs are not well-suited for \textit{multi-step reasoning}---the ability to decompose and solve complex problems through a series of logically related sub-problems \citep{anderson1986learning,schmidhuber90composition,Chollet2019m,LeCunMI22,anderson1993problem}. Despite the remarkable recent progress in ANN research \citep{lecun2015deep,schmidhuber2015deep}, these so-called classic debates persist today.\looseness=-1

In this Perspective, we describe a common framework for training ANNs to overcome the classic challenges, unifying recent work on systematic generalization, catastrophic forgetting, few-shot learning, and multi-step reasoning.
This framework uses metalearning to address what we call the Problem of Incentive and Practice (PIP)---the problem that traditional ANN training is not incentivized to overcome the key weaknesses related to classic challenges, nor does traditional training provide sufficient opportunities to practice improving on these weaknesses.
We discuss how these methods have helped to narrow the gap with human abilities,
how the PIP may be partially addressed in modern large language models trained on internet-scale text, and how these methods relate to broader efforts in using metalearning to better understand human learning \citep{griffiths2020understanding,griffiths2019doing,binz2023meta,ong2024probabilistic,marinescu2024distilling} and development \citep{nussenbaum2024understanding, nussenbaum2024meta,russin2024human}.

\begin{mybox}[boxrule=0pt, parbox=false, opacityframe=0, label=box:glossary]{Glossary}
\begin{itemize}

\item \textbf{Connectionist models:} computational models of cognition and information processing, consisting of (highly simplified) neuron-like processing units and their interconnections. Commonly called artificial neural networks in modern literature.
\item \textbf{Catastrophic forgetting:} forgetting of previously learned material upon learning new material.

\item \textbf{Continual learning:} learning new tasks, one after another, in a sequential manner.

\item \textbf{Few-shot learning:} learning a new concept from just a few examples.

\item \textbf{Metalearning:} 
a learning framework in which a series of learning episodes (called meta-training data) provide opportunity for a learning algorithm (called a metalearner) to practice and improve, i.e., to become a better learner of new episodes.

\item \textbf{Multi-step reasoning:} solving a complex problem by identifying a sequence of simpler sub-problems that lead to a solution.

\item \textbf{Prior:} assumptions about a problem that guide a learner to favor certain solutions.

\item \textbf{Sequence-processing artificial neural networks:} a type of artificial neural network model designed to handle a sequence of inputs.
\item \textbf{Systematic generalization:} the ability to understand or produce novel combinations of familiar components

\end{itemize}
\end{mybox}

\section{Classic Challenges for Artificial Neural Networks}
\label{sec:critiques}
Here we briefly review four classic challenges to ANNs based on comparison to human cognitive capabilities (see Fig.~\ref{fig:classic_challenges}).

\subsection{Systematic generalization}
\label{sec:sys_gen}
People are adept at combining new concepts with existing concepts in rule-like ways. For instance (as illustrated in Fig.~\ref{fig:classic_challenges}a), once a child learns how to ``skip'', they can effortlessly understand how to ``skip twice'' or ``skip all the way to the door.'' These abilities have been characterized as \emph{systematic compositionality}, or the algebraic ability to understand and produce novel combinations of known components.

Over 35 years ago, a seminal paper by Fodor and Pylyshyn \citep{fodor1988connectionism} argued that because ANNs lack this kind of systematicity, they are not a viable model of the human mind, kicking off a lively debate in the late 1980s and 1990s \citep{Smolensky1988, fodor1991connectionism,chalmers1990fodor,hadley1994systematicity,hadley1994systematicity2,niklasson1994being}. We refer to ref.~\citep{frank2009connectionist} for a review on this debate prior to modern deep learning, and ref.~\citep{Russin2024Chapter} for a review until the present.
Several tasks have been proposed as concrete instances of this debate\citep{marcus1998rethinking,marcus2001algebraic,alhama2019review}.
While some of these tasks allow for problem-specific engineering solutions\citep{kurtz2025simple}, the broader challenge is to develop a general framework applicable to any instance of this problem class.

The capabilities of ANNs have advanced dramatically in the last 35 years, and thus recent work has revived the classic debate, proposing new benchmark challenges on systematicity (focusing on operations over sequences, function composition, visual question answering, etc.) and studying whether new architectures show improved capabilities for systematic generalization \citep{lake2017generalization,livska2018memorize,bahdanau2019closure,hupkes2019compositionality,kim2020cogs,keysers2020measuring,csordas2021devil,csordas2021ndr,ctl2022,mccoy2023embers}. Each of these studies points to continued challenges in making systematic generalizations, unless standard architectures are augmented with symbolic machinery or special architectures (e.g., refs.~\citep{ChenLYSZ20,Nye2020}). Thus, even with recent progress in ANNs, the systematicity debate continues.\looseness=-1

\subsection{Continual learning without catastrophic forgetting}
\label{sec:catastrophic_forgetting}
People can typically learn one task and then another task afterwards, without a catastrophic drop in performance on the first task. But when ANNs learn one task after another, without reminders about the earlier tasks
(i.e., data from the previous task becomes inaccessible; the so-called \textit{continual learning} setting---which corresponds to a sequential version of multi-task learning\citep{caruana1997multitask}), there can be a catastrophic drop in performance on the earlier tasks (almost as if the previous tasks were never seen before).
Thus, the problem of \textit{catastrophic forgetting} has been another long-standing challenge of ANNs \citep{mccloskey1989catastrophic,ratcliff1990connectionist,french1991using}.
Fig.~\ref{fig:classic_challenges}b shows a case of the two-task continual image classification setting as an illustration (in fact, two tasks are enough to observe catastrophic forgetting), using example images from the Mini-ImageNet \citep{VinyalsBLKW16,RaviL17,deng2009imagenet} and Omniglot \citep{lake2015human,lake2013one} datasets for Task 1 and 2, respectively.
For further technical details defining variations of continual learning settings in machine learning, we refer to refs.~\citep{hsu2018re,van2019three}.

Human learners can accumulate new skills throughout their lifetime through continual learning, with a more gradual forgetting than the catastrophic forgetting seen in ANNs
\citep{french1999catastrophic,mcclelland1995there}. A common approach is to allow an external memory and replay from previous tasks\citep{robins1995catastrophic}, which, while consistent with certain neural hypotheses \cite{mcclelland1995there}, essentially sidesteps the challenge of achieving continual learning in a single, self-contained ANN.
Although a wide variety of additional algorithms have been proposed to overcome catastrophic forgetting (see ref.~\citep{wang2024comprehensive} for an overview),
continual learning remains a challenge for ANNs.

\subsection{Few-shot learning}
\label{sec:few_shot_learning}
People can learn a new concept from just one or a few examples, an ability known as \textit{few-shot learning} \citep{MillerMV00,FeiFei03,lake2011one,lake2015human,fei2006one}.
For instance (see Fig.~\ref{fig:classic_challenges}c for illustration), a person may learn enough from just one instance of a `Segway' to grasp the boundaries of the concept, allowing them to recognize other members of the class amongst similar-looking objects as distractors.
These capabilities manifest in learning
new visual categories or new faces \citep{carey1980development,biederman1987recognition} as well as in word learning during language acquisition \citep{carey1978child,carey1978acquiring,Bloom2000}.

This sample-efficient learning contrasts with the data hungry learning that usually characterizes ANNs \citep{geman1992neural}.
For example, the best neural language models learn to process language (amonst other abilities) from input with a word count in the trillions, while children learn language from input with a word count in the millions \citep{frank2023bridging}.

\subsection{Multi-step reasoning}
\label{sec:multi_step_reasoning}
People can find a solution to a hard problem by logically decomposing it into a series of simpler sub-problems or steps which can be more easily solved individually.
This ability is characterized as \textit{multi-step reasoning}.
For instance, Fig.~\ref{fig:classic_challenges}d shows an example of a verbal logic question that is difficult to answer immediately after reading the problem statement, but can be solved by enumerating and eliminating possible cases step-by-step.

Since reasoning has traditionally been associated to the capacity for language and symbolic processing (e.g., ref.~\citep{johnson1983mental})---although recent work has argued that humans' capability for language processing and reasoning are quite distinct (e.g., ref.~\citep{fedorenko2016language})---the general critique on the ability to reason has been tightly connected to the systematicity debate (see above).
As a result, rule-based approaches have been considered better suited for reasoning tasks than connectionist models \citep{smith1992case,sun1995robust,browne2001connectionist}.
For example, Newell \citep{newell1990unified} speculated that low-level tasks such as object recognition to be well modeled by ANNs, while higher-level reasoning and logical processing would require rule-based processing.

\section{Overcoming Classic Challenges through Practice}
\label{sec:main}
\subsection{The problem of incentive and practice}
\label{sec:ppa}
What is holding artificial neural networks back from succeeding on these classic challenges?
This Perspective highlights an overlooked issue: the target behaviour expected from the system is not sufficiently well-represented in the objective function that the network seeks to optimize.
Consequently, the ANN is not optimized for the problem it is ultimately expected to solve, and has neither sufficient incentives nor opportunities to practice in improving their behaviour in the relevant way.
We call this the Problem of Incentive and Practice (PIP).\looseness=-1

\begin{figure*}[t]
    \begin{center}
        \includegraphics[width=0.7\columnwidth]{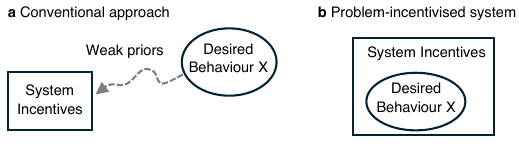}
        \caption{The Problem of Incentive and Practice (PIP) as related to (\textbf{a}) conventional systems and (\textbf{b}) modern systems capable of overcoming the classic challenges (e.g, metalearning systems). \textbf{a:} The Desired behaviour X is unknown/external to the system, except potentially through priors that only indirectly relate to the true problem and objective we ultimately care about.
        \textbf{b:} Metalearning (or the use of appropriate training examples) internalizes the Desired behaviour X into the system's objectives.
        }
        \label{fig:problem_awareness}
    \end{center}
\end{figure*}

The standard continual learning setting is illustrative of this issue. There, an ANN is trained on multiple tasks sequentially, with access to the data from only one task at a time.
The ultimate goal of continual learning is to perform well on all the tasks after having seen all of them.
However, the usual training objective considers performance on only the current task.
Under this objective, the system, or its learning algorithm, is never incentivized to perform well on anything but the current task. From the system's viewpoint, forgetting is entirely compatiable with its specified 
objective of performing well on each task at a time.
The core concept of continual learning lies  outside the scope of the system's learning process, as depicted in Fig.~\ref{fig:problem_awareness}a.

A conventional approach to mitigating weaknesses in ANNs is to inject \textit{priors} or modify the architecture in ways relevant to the desired behaviour into the system, e.g., in continual learning, there are methods that introduce auxiliary terms in the objective to protect previously learned weights \citep{kirkpatrick2017overcoming,ZenkePG17,wang2024comprehensive}.
Often, however, these hand-crafted strategies are indirect or ineffective translations of the desired objective by the human system designer (e.g., for continual learning, an auxiliary loss that optimizes the system to preserve previously learned weights may eventually lead to reduced forgetting, but it does not directly optimize for reduced forgetting; in practice, empirical findings show that such methods are not fully effective \citep{hsu2018re}).

The other challenges (outlined in the previous section) are analogous: in the case of systematicity, conventional ANNs have no direct incentive to ``produce novel combinations of known components''; similarly, they have no incentive to ``learn as much as possible from a few examples'' or to ``reason in multiple steps''.
These observations highlight how a lack of incentives and practice has led researchers to underestimate the abilities of ANNs.
Below, we describe how metalearning (or the use of appropriate training examples, in the case of multi-step reasoning) can be used to formulate the target behaviours within the system incentives (Fig.~\ref{fig:problem_awareness}b), allowing a solution to emerge in a more autonomous, data-driven, practice-based way.

\subsection{Metalearning provides incentive and practice}
\label{sec:common_framework}

Recent work has aimed to overcome classic challenges of ANNs through a metalearning framework that combines powerful \textit{sequence-processing artificial neural networks} with judiciously constructed meta-training sequences and a meta-objective function that reflect the core objectives of the challenge to be addressed.

\begin{figure*}[t]
    \begin{center}
        \includegraphics[width=1.0\columnwidth]{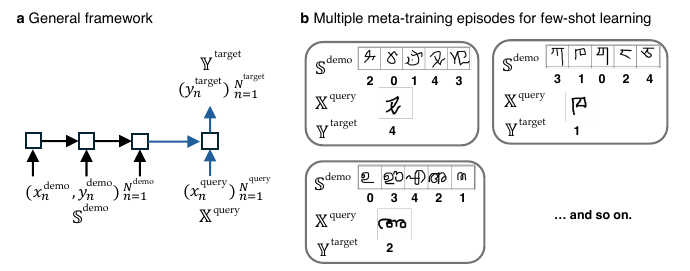}
        \caption{Applying metalearning to address classic challenges for artificial neural networks.
        \textbf{A:} To address the PIP, a metalearning ANN learns, or ``practices'', producing target behaviour $\mathbb{Y}^\text{target}$ corresponding to some challenging query $\mathbb{X}^\text{query}$ after observing some demonstration sequence $\mathbb{S}^\text{demo}$.
        The (meta-)objective provides the model with an ``incentive'' to improve its output in light of expected behaviour.
        \textbf{B:} Three meta-training episodes are illustrated for the few-shot learning case. 
        A single example episode for each challenge can be found in Fig.~\ref{fig:classic_challenges}.
        }
        \label{fig:general_framework}
    \end{center}
\end{figure*}

\paragraph{Metalearning as sequence-processing.}
Standard learning involves training using a fixed learning algorithm on a fixed task given a fixed training dataset. In contrast, metalearning involves training an adaptable learning algorithm capable of learning new tasks given a dynamic sequence of datasets (called episodes). In metalearning, each training episode has  a specific set of training examples (called a demo set; see Fig.~\ref{fig:general_framework}b for three example episodes) and test examples (a query set). As highlighted in this Perspective, with regards to applying metalearning to ANNs, the goal is to find an ANN (by finding its weights) that is capable of handling new problems by reading the demo set into memory, and then, handling queries in the context of that particular demo set\citep{schmidhuber1987evolutionary}.

By treating the demo set in each episode as a sequence of observations, we can parameterize the metalearner by a sequence-processing ANN \citep{cotter1990fixed,cotter1991learning,younger1999fixed,hochreiter2001learning} (see Fig.~\ref{fig:general_framework}a and Fig.~\ref{fig:general_framework}b for illustration).
Such Metalearning Sequence Learners (MSL) are not new: seminal work has discussed the possibility of parameterizing learning dynamics using sequence-processing recurrent neural networks (RNNs) \citep{cotter1990fixed,cotter1991learning,younger1999fixed}, which draw older inspiration from refs.~\citep{rich1964vacuum,white1990new} and inspired a successful implementation by Hochreiter and colleagues \citep{hochreiter2001learning}.
MSL has been revived in the deep learning era of the last decade \citep{bosc2015learning,SantoroBBWL16,duan2016rl,wang2016learning,munkhdalai2017meta,mishra2018a}, and is now often called in-context learning in relation to large language models \citep{gpt3} (e.g., refs.~\citep{XieRL022,Garg22,RaventosPCG23,panwar2023context,von2022transformers,dai2022can,akyurek2022learning,MinLHALHZ22}). In principle, any general-purpose sequence-processing network can be used, e.g., Lake and Baroni \citep{lake2023human} and Irie and colleagues \citep{irie2023automating} respectively use the quadratic \citep{trafo} and linear \citep{schmidhuber1992learning,katharopoulos2020transformers,schlag2021linear,irie2021going} variants of the now popular transformer; although older implementations by Hochreiter and colleagues \citep{hochreiter2001learning} and Santoro and colleagues \citep{Santoro2016} use long short-term memory (LSTM) \citep{Hochreiter1997} and memory augmented RNNs \citep{Graves2016}, respectively.

\paragraph{Meta-training sequences and meta-objective function.}
Addressing the weaknesses of ANNs through this framework involves constructing meta-training  tasks (episodes) and the meta-objective function that accurately reflect the ``problem statement'' and ``objective'' of the classic challenge. This can usually be expressed through three components: (1) the system observes demonstrations (demo examples), (2) the system receives query inputs corresponding to a challenging generalization task (e.g., requiring some nontrivial generalization from the study examples), and finally, (3) as a response to the queries, the system produces outputs which are compared to the desired behaviour. By generating many such meta-training episodes, an ANN can be more directly optimized for the desired behaviour by practicing and improving its skills on these episodes.

More formally,
we define a meta-training episode as a set of three components (see Fig.~\ref{fig:general_framework}a for an illustration): a demonstration sequence $\mathbb{S}^\text{demo}$,
 query inputs  $\mathbb{X}^\text{query}$, and target  behaviour $\mathbb{Y}^\text{target}$.
The demonstration sequence can be generically denoted as a sequence of ``x-y''
(input-output) pairs, $\mathbb{S}^\text{demo}=(\vx^\text{demo}_n, \vy^\text{demo}_n)^{N^\text{demo}}_{n=1}$,
where the specific formats of ``x'' and ``y'' depend the problem at hand (as we'll see later in `\hyperref[sec:app_systematicity]{Overcoming classic challenges with metalearning}' section).
Similarly, the query and target can be generically described as $\mathbb{X}^\text{query}=(\vx^\text{query}_n)^{N^\text{query}}_{n=1}$ and $\mathbb{Y}^\text{target}=(\vy^\text{target}_n)^{N^\text{target}}_{n=1}$, which are sequences or sets of tokens representing some input instruction (of a challenging task) ``x'' and expected behaviour ``y'', respectively, used to evaluate of the system.
$N^\text{demo}$, $N^\text{query}$, and $N^\text{target}$ are positive integers denoting the number of elements in the respective sequences/sets.

The metalearner is optimized to take the demonstration sequence and query input as simultaneous input, and then respond to the query by outputting the right target behaviour.
That is, by denoting the parameter set of the metalearner as $\theta$, we optimize $\theta$ to maximize the probability
$p_{\theta}(\mathbb{Y}^\text{target}|\mathbb{S}^\text{demo}, \mathbb{X}^\text{query})$.
To address PIP,  this meta-objective provides an explicit incentive to the system to improve on the target behaviour, and meta-training across a set of many episodes provide it with opportunities to practice improving on its weaknesses.

\paragraph{Deep Learning Ethos.} Failures of prior approaches to address systematic compositionality and continual learning can be attributed to the difficulty for human engineers to manually designing learning algorithms that give rise to such behaviours (see the previous section).
The complexity of designing such algorithms has historical echoes to that of writing programs and hand-crafting features capable of distinguishing arbitrary images of dogs from those of cats---a domain where the data-driven approach of deep learning has proven more successful, especially when all of the modules are informed about and optimized for the objective at hand (i.e., end-to-end learning). 
Here the metalearning approach follows the same ethos: it aims to discover learning algorithms from data, thereby replacing the manual engineering of such algorithms.

This deep learning ethos is particularly prominent in MSL compared to alternative metalearning approaches, such as model-agnostic metalearning (MAML) \citep{FinnAL17,FinnL18}. While MAML only metalearns a setting of the initial model parameters \citep{JavedW19,BeaulieuFMLSCC20,ConklinWST20}, MSL metalearns the entire learning dynamics in the parameterized sequence-processing ANNs, which often shows better performance as long as the sequence-processing model (e.g., transformer) is powerful enough\citep{mishra2018a,lake2023human,irie2023automating,lee2023recasting,vettoruzzo2024learning}.

\subsection{Overcoming classic challenges with metalearning}
\label{sec:app_systematicity}

\paragraph{Systematic generalization.}
Lake and Baroni's \citep{lake2023human} approach to optimizing ANNs for making stronger systematic generalizations can be described as follows. We use the task in Fig.~\ref{fig:classic_challenges}a as an illustrative example of what a meta-training episode would look like. For a given episode:
\begin{enumerate}
\item The demonstration sequence $\mathbb{S}^\text{demo}$ consists of input primitives paired with their output meanings (e.g., `skip $\rightarrow$ A' and `jump $\rightarrow$ B`, where `A' and `B' are symbols representing the actual behaviours corresponding to `skip' and `jump', and `$\rightarrow$' is used to denote the input-output boundary), as well as demonstrations of more complex, compositional inputs and their outputs (e.g., `jump twice $\rightarrow$ B B').
\item The query input $\mathbb{X}^\text{query}$ contains examples of a novel composition (e.g., `skip twice') to which we want the model to produce the right output.
\item The target $\mathbb{Y}^\text{target}$ contains the expected behaviour corresponding to the query (here, `A A' for the query example above).
\end{enumerate}
Across many episodes, this framework explicitly optimizes the system to generalize to unseen compositions.
Crucially, to construct meta-training examples, the mapping from the primitive to its meaning symbol is randomized for each sequence. For example, if `jump' is defined as `B' in a meta-training sequence, `jump twice' corresponds to `B B', while in another sequence, if `jump' is defined as `C`, `jump twice' should translate to `C C'. This way, the model learns-to-learn the meaning of primitive words by observing their definition provided in the demonstration that is specific to the sequence, preventing the system from just memorizing input-output patterns in its weights.

Regarding the model architecture, Lake and Baroni \citep{lake2023human} use the encoder-decoder transformer \citep{trafo} by feeding the demonstration and query sequences to the encoder, and produce the model's output behaviour using the decoder.
This is an engineering choice; conceptually, one could also use the decoder-only autoregressive transformer instead. Our illustration in Fig.~\ref{fig:general_framework}a shows a generic autoregressive model, without losing the essence of the methodology.

Using this framework, Lake and Baroni \citep{lake2023human} achieved high accuracy in systematic generalization tasks from two popular machine learning benchmarks, SCAN \citep{lake2017generalization} and COGS \citep{kim2020cogs}, as well as in a few-shot instruction-learning task (used to test both humans and machines side-by-side), where the metalearning system was shown to achieve human-like systematic generalization.

\paragraph{Continual learning without catastrophic forgetting.}
Irie and colleagues' \citep{irie2023automating} method to overcome catastrophic forgetting can be obtained as a special case of the framework above as follows.
As an illustrative example, we use the 2-task continual image classification example of Fig.~\ref{fig:classic_challenges}b; we refer to the two tasks as Task 1 and 2, respectively.

\begin{enumerate}
\item The demonstration sequence $\mathbb{S}^\text{demo}$ consists of image-label example pairs constructed by concatenating multiple sub-sequences, each containing examples from a single task. For instance, in the 2-task example, there are two sub-sequences; the first sequence only contains examples from Task 1, and the second one containing examples from Task 2 only.\looseness=-1
\item The query input $\mathbb{X}^\text{query}$ (which is a set of tokens, as the exact order does not matter here) contains unseen images to be classified from all the tasks. In the 2-task example, this corresponds to images from both Task 1 and 2.
\item The target $\mathbb{Y}^\text{target}$ contains the corresponding labels for the query images (the exact order does not matter as long as they are associated to the corresponding query images).
\end{enumerate}
In this formulation, the objective function includes the evaluation of the model performance on all the tasks; this explicitly penalizes the model for forgetting the previous task, thereby teaching the system the core problem of continual learning.

Importantly, Irie and colleagues \citep{irie2023automating} used a variant of linear transformers \citep{IrieSCS22,irie2023practical} that only maintains a fixed-size state as its memory, like standard RNNs, which is updated every time an input is fed to the model. Thus, the model does not automatically store data from previous tasks, it must metalearn what to store through incentives and practice. This contrasts with regular transformers that automatically stores previous tasks in its memory, a potential violation of the continual learning setting. Ref.'s \citep{irie2023automating} formulation achieved successful continual learning without catastrophic forgetting on the classic machine learning benchmark, Split-MNIST\citep{SrivastavaMKGS13,ZenkePG17}, where previous hand-crafted learning algorithms have not been successful.

\paragraph{Few-shot classification.}
Few-shot learning is another classic challenge that can be approached within the same framework, using MSL to address the PIP.
An episode of few-shot image classification can be stated as follows (see Fig.~\ref{fig:general_framework}b for an example of few-shot classification of handwritten characters \citep{lake2015human}).

\begin{enumerate}
\item The demonstration sequence $\mathbb{S}^\text{demo}$ consisting of a $K$ image-label example pairs for each class (this is effectively a set because the order does not matter), where $K$ is a positive integer denoting the number of training examples available per class in the $K$-shot learning scenario.
\item The query input $\mathbb{X}^\text{query}$ contains unseen images to be classified.
\item The target $\mathbb{Y}^\text{target}$ contains the labels corresponding to the query images (the order does not matter as long as they are associated to the correct query images). These labels can change for every episode.
\end{enumerate}

This framework meta-trains the system to make predictions only after observing a few training examples; thereby, it explicitly incentivizes the system to do its best to learn new concepts from just a few examples. This application of the MSL framework predates the other applications above, and the current popularity of in-context learning.
For instance, Santoro and colleagues \citep{SantoroBBWL16} (2016) trained a memory-augmented RNN \citep{graves2016hybrid} for in-context few-shot image classification; as one of the contemporary methods that implemented memory-based metalearning \citep{VinyalsBLKW16,RaviL17,SnellSZ17},
and it was later extended to the transformer-family by Mishra and colleagues \citep{mishra2018a} (2018).
Earlier, Hochreiter and colleagues \citep{hochreiter2001learning} (2001) trained LSTM RNNs for in-context few-shot regression, achieving improved sample efficiency compared to the regular gradient-descent learning algorithm.\looseness=-1

Similar to the continual learning case above, these metalearned learning algorithms have been empirically shown to achieve few-shot learning performance on popular benchmarks, including Omniglot \citep{lake2015human} and Mini-ImageNet \citep{VinyalsBLKW16,RaviL17}, that is often beyond the reach of hand-crafted learning algorithms.

\paragraph{Multi-step reasoning.}

Multi-step reasoning is another challenge that can be analyzed through this framework. Although not currently studied in an explicitly metalearning setting (the demonstration sequence is often empty), it can still be positioned with the general idea of addressing the PIP (see Fig.~\ref{fig:classic_challenges}d). We use the task from Fig.~\ref{fig:classic_challenges}d as a representative example; the setting can be formulated as follows.

\begin{enumerate}
\item The demonstration sequence $\mathbb{S}^\text{demo}$ is typically empty. More generally, $\mathbb{S}^\text{demo}$ could contain examples of reasoning tasks, together with specific intermediate steps for solving them.
\item The query input  $\mathbb{X}^\text{query}$ is the task statement (i.e., the text in the box in Fig.~\ref{fig:classic_challenges}d).
\item The target $\mathbb{Y}^\text{target}$ is the entire response text containing the reasoning steps and the answer.
\end{enumerate}

Although the reasoning capabilities of LLMs are still limited (e.g., ref.~\citep{lampinen2024language}), multi-step reasoning can be facilitated in certain settings with explicit prompts that encourage step-by-step processing \citep{nye2021show,cobbe2021training,KojimaGRMI22,wei2022emergent,Wei0SBIXCLZ22}.
These abilities are related to the models' training incentives and opportunities for practice; the internet-scale training text contains many examples of such multi-step reasoning, which explicitly incentivize the model to produce such intermediate reasoning steps during training.
In practice, reasoning can be further strengthened with more explicit incentives and practice: for instance, curating datasets to provide more precise supervision regarding intermediate steps can improve reasoning\citep{RajaniMXS19,lightman2023let,kirchner2024prover,ZelikmanWMG22}.
Moreover, recent work \citep{gandhi2024stream} shows how training an LLM on synthetically generated episodes of step-by-step reasoning can improve performance on related problems, consistent with the incentive and practice framework articulated here.
Overall, leveraging appropriate training datasets that incentivize multi-step reasoning currently stands out as the most prominent approach to addressing the PIP and building artificial neural networks with this capability. We further discuss relationships between LLMs and the PIP in `\hyperref[sec:llms]{Capabilities of large-scale AI systems}' section below.

\section{Discussion}
\label{sec:discussion}
In this Perspective, we discussed recent advances in ANNs that use metalearning to find data-driven solutions to address their weaknesses. This family of methods, while developed for different aims and by different groups, can be viewed through a common lens: providing ANNs with incentive and practice to overcome classic obstacles. When trained in this way, ANNs become more proficient in understanding novel combinations of known components, learning without forgetting, learning from just one or a few examples, or reasoning through a series of logical steps. 
In this sense, metalearning and advances in sequence processing have pushed the limits of what ANNs can do, challenging classic skeptical arguments about their capabilities, without requiring any changes to the basic architectures used in contemporary AI. This raises a number of natural questions about the limitations of this framework, the connection to large-scale AI systems, and the implications for human cognitive development, each discussed in the sections below.

\subsection{Limitations and outstanding questions}
The metalearning framework for addressing PIP still has important limitations.
First, the problem of interest must be suitably posed using the three-part formulation consisting of demonstration, query, and target (see `\hyperref[sec:common_framework]{Metalearning provides incentive and practice}' section above). We showed how this framework can be quite general, but it remains to be seen whether it covers the full range of behaviours we expect from cognitive models, e.g., in other types of abstraction and reasoning problems \citep{ShanahanM22} or in the AI safety and alignment domain \citep{hubinger2019risks,Ouyang0JAWMZASR22}.
One candidate for the future application of this approach is the Abstraction and Reasoning Challenge (ARC) by Chollet \citep{chollet2019measure}. The problem formulation of ARC naturally fits the metalearning framework for few-shot learning: answering a new question (query) given a few examples of a novel task (demonstrations).
We leave the practical experiments as an important avenue for the future, although the limited number of training problems provided in the ARC challenge may make generalization challenging, which leads us to the next limitation.

Second, metalearning relies on the well-known in-distribution generalization capabilities of ANNs; conversely, it is inherently limited by ANNs' out-of-distribution generalization abilities. For example, a model that has been meta-trained to decompose some problems in multiple reasoning steps will likely not generalize to highly different problems (`multi-step reasoning' can be replaced by `continual learning' or `compositional generalization' to obtain an analogous statement describing limitations in other challenges). It remains an open challenge to build a meta-meta-learning framework to support generalization of the high-level skills for compositional learning, continual learning, few-shot learning, or multi-step reasoning abilities, across different problem settings.
This may require meta-training increasingly large models on increasingly varied problems, such as ref.'s \citep{PFNNature} scaling up of metalearning.

Last, metalearning relies on a process for generating meta-training episodes, which usually requires human ingenuity and hand-engineering.
The case of large language models (LLMs) may be an exception in that it relies on implicit episodes of experience, which we discuss in the next section.\looseness=-1

\subsection{Capabilities of large-scale AI systems}
\label{sec:llms}
Here we highlight the relevance of PIP in the current era of foundation models trained on vast amounts of internet-scale data. We argue that, for some of their capabilities, PIP offers conceptual explanations for their origin, while for some of their limitations, PIP provides insights into potential solutions to overcome them.

Language modelling can be interpreted as sequence processing with error feedback, where the correct label to be predicted is fed to the model with a delay of one time step (as in Fig.~\ref{fig:general_framework}a; see also refs.~\citep{gpt3,hochreiter2001learning}).
As discussed above, this introduces a natural incentive for learning as soon as each input becomes available, which is the essence of few-shot learning, and thus auto-regressive next-token prediction can provide a few-shot learning incentive \citep{gpt3}. Language modeling can also provide the right incentives and practice for other abilities. For multi-step reasoning, text from the internet used to train LLMs naturally contains step-by-step problem-solving examples, especially in pedagogical contexts. For continual learning, it is conceivable that similar topics or tasks re-occur within the same context window, incentivizing the model to remember previous materials, which gives rise to in-context continual learning-like objectives akin to the one discussed in this work.
For systematic generalization, text that introduces a novel concept and then mentions that concept again (especially in a novel, compositional context within the same context window) provides a real-world opportunity for compositional learning, although additional engineering (or more concentrated meta-learning) may be needed to address infrequent compositions \citep{conwell2022testing,betker2023improving} and bidirectional relationships \citep{BerglundTKBSKE24}.

Overall, some of the most impressive capabilities in modern LLMs are likely related to addressing the PIP through the structure of natural training data. Moreover, standard LLM training and metalearning can be highly complementary; that is, adding more explicit metalearning incentives to LLM training can lead to additional abilities and stronger performance \citep{WangMinnow}.
Still, the principles for achieving human-like performance at realistic human-like learning scales are not yet clear: modern LLMs are trained on internet-scale text with a word count in the trillions, far beyond the millions that children need to learn language \citep{frank2023bridging}.

\subsection{Outlook on human learning and development}

The experience-driven perspective inherent to metalearning raises important questions related to human learning and development.
Does cognitive development, paired with the natural environments in which it occurs, provide incentives to make challenging generalizations and opportunities to practice those generalizations? Do natural environments, and the natural incentives of development, provide a more realistic but still effective replacement for synthetic meta-training episodes of experience \cite{Smith2022,Chan2022d}?
As discussed in recent work \cite{lake2023human,russin2024human} and in the above discussion of LLMs, one possibility is that natural environments provide humans with scenarios where there is a pressure to learn new concepts quickly (few-shot learning) and use them compositionally right away (systematic generalization). Similar pressures could also show a continual learner how previous tasks can end up being relevant during later learning, and thus there's a process of learning to learn whereby learning skills improve over development.

There are developmental findings consistent with this idea. Children become better word learners over development \citep{Bergelson2020}, and childrens' experience has been linked to improving abilities for few-shot learning \citep{smith-etal02}. Relating to systematic generalization, pre-school children, but not infants, seem to make meaningful inferences about visual function composition \citep{Piantadosi2016b,piantadosi2018limits}, a change that could be linked to maturity and/or experience.
For multi-step reasoning, formal education provides humans with opportunities to learn and practice these skills.
Similarly, memory can also be honed through experience and undergoes key changes in development \citep{coffman2019relating}, and potentially, if this framework provides the right lens for development, in reaction to environmental and developmental needs. Finally, it has been noted that the characteristics of early environments (e.g., reward availability, volatility, and controllability) shapes not only the knowledge learned but the future reward learning mechanism itself  \cite{nussenbaum2024meta,nussenbaum2024understanding}, the signature of metalearning.

However, we are unaware of successful metalearning models trained on episodes that emulate human development environments, statistics, and data quantities (e.g., ref.~\citep{Vong2024Science}), a crucial step for evaluating this framework as a developmental approach.
Conversely, if replicating human-like development could help artificial neural network models acquire more human-like abilities, would training models on more natural input streams (e.g., headmounted video form children) lead to additional AI advances (cf.~ref.~\citep{orhan2024learning})?
Such pursuits would also raise an even broader question: what is the computational basis of various stages of development, from evolutionary to educational, and how to capture them in future computational models of development (cf.~ref.~\citep{shuvaev2024encoding})?

Beyond incentives and opportunities to practice, people are also influenced by more explicit metacognitive knowledge about the goals they have and the problems they seek to solve.
For example, if human participants are told they are in a continual learning setting (e.g., Fig.~\ref{fig:classic_challenges}b) such that they will learn a series of tasks, and that previously learned tasks will show up in later tests, that would surely alter their strategies and behaviour. Standard ANN models, if trained naively on a sequence of tasks, would lack awareness of the full nature of the problem they will be evaluated, that is, they suffer from a problem of problem awareness.
Integrating this kind of metacognitive knowledge into ANN-based cognitive models is another avenue for future work, distinct yet related to the more implicit incentives discussed here, that recent advances in AI could also fruitfully contribute to.\looseness=-1

\section*{Acknowledgement}

Kazuki Irie thanks Sam Gershman for valuable feedback on an earlier version of this work, and R\'obert Csord\'as and Imanol Schlag for introducing him to the systematicity challenge a few years ago at the Swiss AI lab IDSIA.

\section*{Competing interests}

The authors declare no competing interests.

\bibliography{references}
\bibliographystyle{naturemag}
\end{document}

%% file: math_commands.tex
\usepackage{amsmath,amsfonts,bm}

{}
{}
{}
{}

\def\eqref#1{equation~\ref{#1}}

\def\1{\bm{1}}

\def\vx{{\bm{x}}}
\def\vy{{\bm{y}}}

\DeclareMathAlphabet{\mathsfit}{\encodingdefault}{\sfdefault}{m}{sl}
\SetMathAlphabet{\mathsfit}{bold}{\encodingdefault}{\sfdefault}{bx}{n}

